\documentclass[conference]{IEEEtran} 

\usepackage[hidelinks,bookmarks=false]{hyperref}

\usepackage{tikz}
\usetikzlibrary{shapes.geometric, arrows, positioning, fit, backgrounds}
\usepackage{graphicx}
\usepackage{booktabs}
\usepackage{tabularx}
\usepackage{array}
\usepackage{cite}
\usepackage{amsmath}
\usepackage{xcolor}
\usepackage{listings}
\usepackage{braket}
\usepackage{algorithm}
\usepackage{algpseudocode}
\usepackage{float}
\usepackage{ragged2e}

\definecolor{lightgray}{gray}{0.95}
\begin{document}

\title{Quantum-Enhanced Forecasting for Deep Reinforcement Learning in Algorithmic Trading }

\author{
Jun-Hao Chen$^1$, Yu-Chien Huang$^1$, Yun-Cheng Tsai$^1$, Samuel Yen-Chi Chen$^2$\\
$^1$National Taiwan Normal University\\
$^2$Wells Fargo, USA\\
\small \texttt{chenjunhao15@ntnu.edu.tw, 41171218H@gapps.ntnu.edu.tw,}\\ \texttt{pecu@ntnu.edu.tw, yen-chi.chen@wellsfargo.com}
}

\setlength{\textwidth}{6.625in}
\maketitle

\begin{abstract}
The convergence of quantum-inspired neural networks and deep reinforcement learning offers a promising avenue for financial trading. We implemented a trading agent for USD/TWD by integrating Quantum Long Short-Term Memory (QLSTM) for short-term trend prediction with Quantum Asynchronous Advantage Actor-Critic (QA3C), a quantum-enhanced variant of the classical A3C. Trained on data from 2000-01-01 to 2025-04-30 (80\% training, 20\% testing), the long-only agent achieves 11.87\% return over around 5 years with 0.92\% max drawdown, outperforming several currency ETFs. We detail state design (QLSTM features and indicators), reward function for trend-following/risk control, and multi-core training. Results show hybrid models yield competitive FX trading performance. Implications include QLSTM's effectiveness for small-profit trades with tight risk and future enhancements. Key hyperparameters: QLSTM sequence length$=$4, QA3C workers$=$8. Limitations: classical quantum simulation and simplified strategy.
\footnote{The views expressed in this article are those of the authors and do not represent the views of Wells Fargo. This article is for informational purposes only. Nothing contained in this article should be construed as investment advice. Wells Fargo makes no express or implied warranties and expressly disclaims all legal, tax, and accounting implications related to this article. }
\end{abstract}


\section{Introduction}

Financial markets generate vast, noisy, and highly non-stationary data streams that challenge traditional modeling techniques. Deep reinforcement learning (DRL) has emerged as a powerful alternative, as it frames trading as a sequential decision-making task. Algorithms such as Asynchronous Advantage Actor-Critic (A3C) learn directly from reward feedback, rather than relying on fixed heuristics~\cite{luong2019applications, samsami2020distributeddeepreinforcementlearning, arabha2024improvingdeepreinforcementlearning}. A3C's parallel actor-learner architecture stabilizes training and has delivered state-of-the-art performance on many sequential control problems, motivating its adoption in algorithmic trading~\cite{li2019deep, zhang2019deep, yang2020deep}.

Yet DRL agents still depend critically on the quality of the state representation they receive. Suppose the raw price series is fed directly to the network. In that case, the agent must simultaneously discover predictive patterns and learn optimal actions, inflating sample complexity and increasing the risk of converging on noisy correlations. A common remedy is to prepend a supervised forecaster — typically a deep Long Short-Term Memory (LSTM) network — that distills historical data into a succinct signal, such as the "probability of an upward move in the next k days ~\cite{sunny2020deep, ghosh2022forecasting}."  The trading agent then focuses on decision-making, effectively turning the forecaster into an expert feature extractor.

Recent advances in quantum machine learning (QML) offer fresh avenues, with models like quantum neural networks (QNNs) and variational quantum algorithms (VQAs) exploiting superposition and entanglement for non-linear mappings in Hilbert space~\cite{Biamonte2017, schuld2015introduction, cerezo2021variational}. This exposes hidden patterns in noisy financial data, as demonstrated in chaotic series and anomaly detection~\cite{dunjko2016quantum, preskill2018quantum, ciliberto2018quantum}.

We combine a Quantum Long Short-Term Memory (QLSTM) network \cite{chen2022QLSTM} for short-horizon trend forecasting with a Quantum Asynchronous Advantage Actor-Critic (QA3C) agent \cite{chen2023asynchronous} to determine when to buy, sell, or hold the USD/TWD currency pair. This architecture simultaneously addresses predictive richness by supplying probabilities for price rises or falls exceeding $\pm1.2\%$ over five days through QLSTM's quantum variational layers, sample-efficient learning by outsourcing pattern discovery to QLSTM, and risk-aware control via a reward function that incorporates penalties and bonuses.

We evaluate the hybrid QLSTM-QA3C agent on 25 years of daily USD/TWD data (from January 1, 2000, to April 30, 2025), reserving the most recent 20\% for out-of-sample testing. Despite using a simplified long-only strategy with fixed position sizing, the agent achieves an 11.87\% total return over the 2020--2025 test window with a maximum drawdown of just 0.92\%, outperforming or matching several major currency ETFs on a risk-adjusted basis. By demonstrating that quantum-enhanced forecasts can materially improve a quantum variant of the classical DRL trading policy (QA3C), this work contributes evidence that quantum machine learning is more than a theoretical curiosity: even today, quantum-inspired models simulated on classical hardware can deliver tangible performance gains in finance. The remainder of the paper details the architecture, training protocol, empirical results, and avenues for extending the approach to multi-asset and fully quantum implementations.


\section{Methodology}
\subsection{Data and Preprocessing}
Daily USD/TWD quotes from January 1, 2000, to April 30, 2025, form the raw corpus. Records are sorted chronologically, de-duplicated, and any row with a missing or zero close price is discarded. We compute four moving averages (MA5, MA10, MA20, MA60) plus a 20-day rolling volatility. Features are normalized via min-max scaling to $[0, 1]$. The series is split temporally: the first 80\% (from 2000-01-01 to 2020-03-31) is reserved for model training, and the final 20\% (from 2020-04-01 to 2025-04-30) is used for out-of-sample testing. 

\subsection{Quantum Long Short-Term Memory (QLSTM)}
Our primary state generator for the subsequent reinforcement learning (RL) model is a Quantum-enhanced Long Short-Term Memory (QLSTM) \cite{chen2022QLSTM} that functions as a price-movement forecaster. Concretely, input features are first normalized and then directly encoded as angle parameters $\theta$ which control single-qubit rotations, such as the $R_y(\theta)$ gate defined as:
\[
R_y(\theta) = \begin{pmatrix} \cos(\theta/2) & -\sin(\theta/2) \\ \sin(\theta/2) & \cos(\theta/2) \end{pmatrix}.
\]
Entangling CNOT layers follow. These quantum circuits, which output expectation values, act as replacements for the classical linear layers within the LSTM's gating mechanisms. The resulting expectation values are passed through standard sigmoid and tanh activation functions. The hybrid cell, therefore, executes a non-linear, high-dimensional feature map while retaining the temporal recurrence of a standard LSTM. The quantum circuits are simulated using the combination of PennyLane and PyTorch library.

The QLSTM consists of a hidden state dimension of 2 followed by a dense-softmax output head to predict the price direction. Inputs are sequences of the last 4 time steps, each comprising six normalized features: open, high, low, close, MA5, and MA10. For supervision, each day, $t$ receives a binary label by looking five days ahead: labeled "up" if the price increases by more than 1.2\%, and "down" if it decreases by more than 1.2\%; intermediate cases are excluded. 
Formally, the model's task is to predict the binary outcome for day $t$ based on the condition $\frac{y(t+5) - y(t)}{y(t)} > \pm 1.2\%$, given the input feature sequence $\{x(t-3), x(t-2), x(t-1), x(t)\}$.
The network is trained on these input-label pairs over the first 80\% of the dataset for 50 epochs using RMSprop (learning rate $= 5e-3$) and weighted cross-entropy loss. Testing accuracy stabilizes around 71.5\%.

For the RL agent, the trained QLSTM is frozen and queried online. The latest 4-day window is normalized, passed through the network, and the resulting two-element softmax vector is appended to the RL state. Crucially, the goal of this supervised learning step is not to achieve high prediction accuracy, but rather to compel the model to extract salient features of upward and downward trends from the raw data, providing useful state information for the subsequent QA3C model.

\begin{table*}[htbp]
\centering
\caption{Observation Features ($s_t$) used by QA3C agent at each trading day $t$}
\begin{tabular}{l|p{12cm}}
\hline
\textbf{Feature} & \textbf{Description} \\
\hline
QLSTM Bullish Probability & QLSTM softmax probability that the price will increase by more than +1.2\% within five days. \\
QLSTM Bearish Probability & QLSTM softmax probability that price will decrease by more than -1.2\% within five days. \\
Cash Ratio & Current cash divided by initial capital. \\
Holdings Ratio & Market value of current holdings divided by initial capital. \\
Unrealized P\&L\% & Percentage profit or loss of the current open position. \\
Average Price Ratio & Difference between current price and average cost, normalized by current price. \\
Price-to-MA20 Deviation & Percentage deviation of current price from its 20-day moving average. \\
Price-to-MA60 Deviation & Percentage deviation of current price from its 60-day moving average. \\
MA5 vs MA20 Gap & Difference between 5-day and 20-day moving averages (MA5 $-$ MA20). \\
20-day Relative Volatility & Rolling 20-day volatility, computed as standard deviation divided by mean ($\frac{\sigma}{\mu}$). \\
\hline
\end{tabular}
\label{tab:QA3C_features}
\end{table*}

\subsection{Quantum Asynchronous Advantage Actor-Critic (QA3C) Agent}
The QA3C module marries a classical asynchronous actor-critic loop with quantum-assisted features. At each trading day $t$, the QA3C agent receives a 10-dimensional observation vector $s_t$, whose components are detailed in Table~\ref{tab:QA3C_features}.

We adopt a simplified discrete action space consisting of only three actions: buy, sell, and hold. The agent is restricted to taking long positions only; short selling is not allowed. If sufficient cash is available, the agent can accumulate multiple long positions through repeated purchases; however, all holdings must be liquidated in a single sell action. The discrete action set is the following:
\[
A = \left\{
\begin{array}{ll}
0: & \text{hold} \\
1: & \text{buy one unit if cash} \geq \text{unit price} \\
2: & \text{sell all}
\end{array}
\right.
\]

The policy/value network $f_{\theta}$ employs a hybrid quantum-classical architecture. The input state $s_t$ is first mapped to an 8-dimensional latent space through a $\tanh$ activated fully-connected layer. This vector then passes through an 8-qubit Variational Quantum Circuit (VQC), whose output is subsequently fed into two separate linear heads to produce the policy $\pi_{\theta}(\cdot\,|\,s_t)$ and the value $V_{\theta}(s_t)$. Multiple workers interact with independent environments, accumulate trajectories of length $n=30$, and asynchronously update the shared parameters with shared Adam ($\mathrm{lr} = 1e-5$, $\gamma = 0.995$, $\beta_{\mathrm{entropy}} = 0.05$). The training and update logic is summarized in Algorithm~\ref{alg:qa3c}.

Our proposed QA3C model demonstrates superior performance and parameter efficiency compared to its classical counterpart. It yields a 0.45\% higher total return on testing sets than the standard A3C, while using only 244 trainable parameters (32 quantum, 212 classical) in stark contrast to the classical model's 3,332. The strategic integration of quantum variational layers makes this significant parameter reduction possible.

Furthermore, while end-to-end training of quantum recurrent models like QLSTM in RL directly is often computationally expensive due to the required backpropagation-through-time, our framework circumvents this bottleneck to gain a significant computational advantage. We achieve this by employing a two-stage approach: first, a QLSTM is pre-trained to act as a feature extractor, and is subsequently frozen. This allows the main QA3C agent to be trained efficiently on these pre-computed features without the need to backpropagate through the quantum recurrent neural network. This computational efficiency allows for a greater model capacity. For instance, we set the hidden layers in the actor and critic networks to a dense size of 8 units to enhance learning performance. As detailed in Table~\ref{Parameter}, this hybrid QA3C architecture is efficient to train and remarkably parameter-efficient, requiring significantly fewer parameters than its purely classical A3C counterpart.

\begin{table}[htbp]
\centering
\caption{Parameter Comparison}
\begin{tabular}{l|c|c|c}
\hline
\textbf{Model} & \textbf{Quantum} & \textbf{Classical} & \textbf{Total} \\
\hline
A3C & 0 & 3,332 & 3,332 \\
QA3C & 32 & 212 & 244 \\
\hline
\end{tabular}\label{Parameter}
\end{table}

\subsection{Reward Function}
The reward function $r_t$ is designed to promote trend-following behavior while enforcing strict risk control. It combines several components, summarized in Table~\ref{tab:reward_components}, and is aggregated as:
$r_t = -0.02 + \delta_{\text{entry}} + \delta_{\text{exit}} + \delta_{\text{holding}}$
\begin{table*}[htbp]
\centering
\small
\caption{Components of the QA3C Reward Function}
\begin{tabular}{l|p{13cm}}
\hline
\textbf{Component} & \textbf{Description} \\
\hline
Time Cost & A constant daily penalty of $-0.02$, encouraging active decision-making instead of idleness. \\
Entry Reward / Penalty & +0.5 reward for entering with the prevailing trend (e.g., price $>$ MA20); $-2.0$ penalty for counter-trend entries. Invalid actions (e.g., purchasing without sufficient funds) incur minor penalties. \\
Exit Reward / Penalty & Profitable sells receive $+10 + 50 \times \mathrm{pnl}\%$; losing sells are penalized with $-2 - 10 \times \mathrm{pnl}\%$, balancing punishment with trade encouragement. \\
Holding Penalty & For unrealized losses, a daily quadratic penalty of $-5 \times (\mathrm{pnl}\%)^2$ discourages deep drawdowns. \\
Reward Clipping & Final $r_t$ is clipped to the range $[-15, +30]$ to stabilize learning gradients. \\
\hline
\end{tabular}
\label{tab:reward_components}
\end{table*}

\subsection{Integration of QLSTM and QA3C}
\begin{figure*}[htbp]
\centering
\small
\begin{tikzpicture}[node distance=0.75cm,
    every node/.style={fill=white, font=\sffamily, align=center},
    arrow/.style={thick, -latex}]

\node (start) [ellipse, fill=gray!20] {Start};
\node (data) [rectangle, rounded corners, minimum width=3cm, minimum height=1cm, text centered, draw=black, below=of start] {Data Collection\\(USD/TWD quotes 2000-2025)};
\node (preprocess) [rectangle, rounded corners, minimum width=3cm, minimum height=1cm, text centered, draw=black, below=of data] {Preprocessing\\(De-duplicate, Compute MAs/Volatility, Normalize, Split 80/20)};
\node (trainqlstm) [rectangle, rounded corners, minimum width=3cm, minimum height=1cm, text centered, draw=black, below=of preprocess] {Train QLSTM\\(Sequences of 4 days, Binary labels ±1.2\% over 5 days, RMSprop, Cross-entropy)};
\node (freezeqlstm) [rectangle, rounded corners, minimum width=3cm, minimum height=1cm, text centered, draw=black, below=of trainqlstm] {Freeze QLSTM\\(Query for predictions)};
\node (statespace) [rectangle, rounded corners, minimum width=3cm, minimum height=1cm, text centered, draw=black, below=of freezeqlstm] {Define State Space\\(QLSTM probs + 8 features: Cash/Holdings/PnL/etc.)};
\node (actions) [rectangle, rounded corners, minimum width=3cm, minimum height=1cm, text centered, draw=black, below=of statespace] {Define Actions\\(Hold, Buy (if cash), Sell all)};
\node (reward) [rectangle, rounded corners, minimum width=3cm, minimum height=1cm, text centered, draw=black, below=of actions] {Define Reward Function\\(Time cost, Entry filter, Exit incentives, Holding discipline, Clipping)};
\node (integration) [rectangle, rounded corners, minimum width=3cm, minimum height=1cm, text centered, draw=black, below=of reward] {Integration QLSTM-QA3C\\(Concat preds to state, Pseudocode loop)};
\node (trainingloop) [rectangle, rounded corners, minimum width=3cm, minimum height=1cm, text centered, draw=black, below=of integration] {Training Loop\\(Parallel workers, Rollouts, Advantages, Updates with shared Adam)};
\node (end) [ellipse, fill=gray!20, below=of trainingloop] {Trained Agent};

\draw [arrow] (start) -- (data);
\draw [arrow] (data) -- (preprocess);
\draw [arrow] (preprocess) -- (trainqlstm);
\draw [arrow] (trainqlstm) -- (freezeqlstm);
\draw [arrow] (freezeqlstm) -- (statespace);
\draw [arrow] (statespace) -- (actions);
\draw [arrow] (actions) -- (reward);
\draw [arrow] (reward) -- (integration);
\draw [arrow] (integration) -- (trainingloop);
\draw [arrow] (trainingloop) -- (end);

\begin{scope}[on background layer]
    \node [fit=(trainqlstm)(freezeqlstm), fill=blue!10, rounded corners, inner sep=5pt, label=above:QLSTM Component] {};
\end{scope}

\begin{scope}[on background layer]
    \node [fit=(statespace)(actions)(reward)(integration)(trainingloop), fill=green!10, rounded corners, inner sep=5pt, label=above:QA3C Agent] {};
\end{scope}

\end{tikzpicture}
\caption{Overview of the QLSTM-QA3C methodology workflow. The flowchart depicts the sequential steps from data handling to model training, emphasizing the integration of quantum-enhanced forecasting with reinforcement learning.}
\label{fig:methodology-flowchart}
\end{figure*}
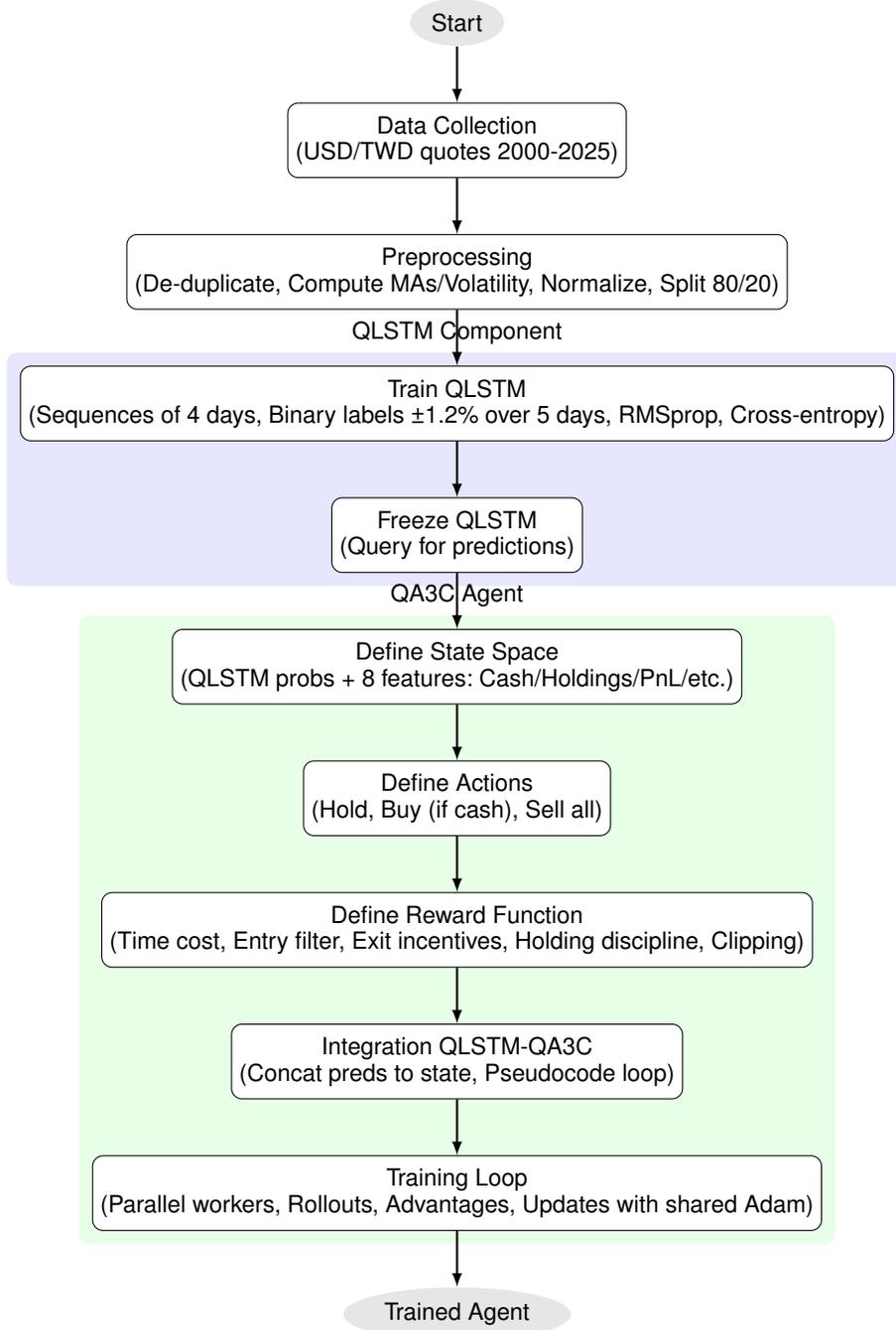

QLSTM predictions are fed as state features to QA3C. 
\begin{algorithm}[htbp]
\caption{QLSTM-QA3C Integration}
\raggedright
\label{alg:integration}
\begin{algorithmic}[1]
\State Initialize QLSTM and QA3C models.
\For {each episode}
    \For {t in test data}
        \State qlstm\_pred = QLSTM(history[t-4:t])
        \State state = concat(qlstm\_pred, features[t])
        \State action = QA3C.policy(state)
        \State next\_state, reward = env.step(action)
        \State Update QA3C with advantage.
    \EndFor
\EndFor
\end{algorithmic}
\end{algorithm}
\begin{algorithm}[htbp]
\caption{QLSTM-QA3C Training Loop}
\raggedright
\label{alg:qa3c}
\begin{algorithmic}[1]
\State \textbf{Input:} historical data $D$, QLSTM forecaster $\mathcal Q$
\State Initialize \textbf{global} network parameters $\theta$, shared Adam optimizer
\For {each worker $w \in \{1,\dots,W\}$ \textbf{in parallel}}
    \State Reset environment; set local parameters $\theta^{(w)} \leftarrow \theta$
    \While {episode $<$ \texttt{max\_ep}}
        \State Clear rollout buffers $\mathcal B \gets \emptyset$
        \For {$t = 1$ \textbf{to} $n$}
            \State Observe price window; 
            \State Obtain QLSTM logits $\bigl(p_\uparrow,p_\downarrow\bigr)=\mathcal Q(\text{window})$
            \State Build state $s_t=[p_\uparrow,p_\downarrow,\text{8 features}]$
            \State Sample action $a_t \sim \pi_{\theta^{(w)}}(\cdot|s_t)$
            \State Execute $a_t$, receive reward $r_t$ and $s_{t+1}$
            \State Append $(s_t,a_t,r_t)$ to $\mathcal B$
            \If {terminal} 
                \State \textbf{break}
            \EndIf
        \EndFor
        \State Compute bootstrapped value $R$:
        \[
        R =
        \begin{cases}
          V_{\theta^{(w)}}(s_{t+1}), & \text{if not terminal} \\
          0, & \text{otherwise}
        \end{cases}
        \]
        
        \For {$(s,a,r)$ in $\mathcal B$ reversed}
            \State $R \leftarrow r + \gamma R$  \Comment{n-step return}
            \State Store advantage $\delta \gets R - V_{\theta^{(w)}}(s)$
            \State Accumulate gradients of loss $L$:
            \[
            L = \delta^2 + \bigl(-\log\pi_{\theta^{(w)}}(a|s)\bigr)\delta - 
                \beta_{\text{entropy}}\,\mathcal{H}\bigl[\pi_{\theta^{(w)}}(\cdot|s)\bigr]
            \]
            
        \EndFor
        \State \textbf{Push} local gradients to global $\theta$; 
        \State \textbf{pull} updated $\theta$ back
    \EndWhile
\EndFor
\end{algorithmic}
\end{algorithm}
To provide a visual overview of the entire methodology, Figure~\ref{fig:methodology-flowchart} illustrates the step-by-step process, from data collection and preprocessing to QLSTM training, state and action definition, reward design, integration of QLSTM predictions into the QA3C state, and the parallel training loop. This flowchart illustrates the hybrid quantum-classical pipeline, where QLSTM handles predictive feature extraction and QA3C focuses on decision-making and policy optimization.

\section{Experiments and Results}

Our agent achieved a 11.87\% total return on the test set with minimal drawdown. Table~\ref{tab:testing-performance} summarizes the results. To put this into context: while the USD appreciated against the TWD during this period, our agent's strategy was not simply to "buy and hold" USD. Instead, it was actively trading – entering and exiting – to compound gains and avoid downturns. The very low drawdown of ~0.92\% suggests that the agent either maintained a cash position during any major unfavorable moves or cut losses before they became substantial. The equity curve (portfolio value over time) in testing shows a steady upward slope with very shallow dips. Each trade contributed a small incremental gain. The agent's style can be characterized as a frequent trading, low-gain, low-loss strategy.

\begin{figure}[!htbp]
\centering
\includegraphics[width=0.49\textwidth]{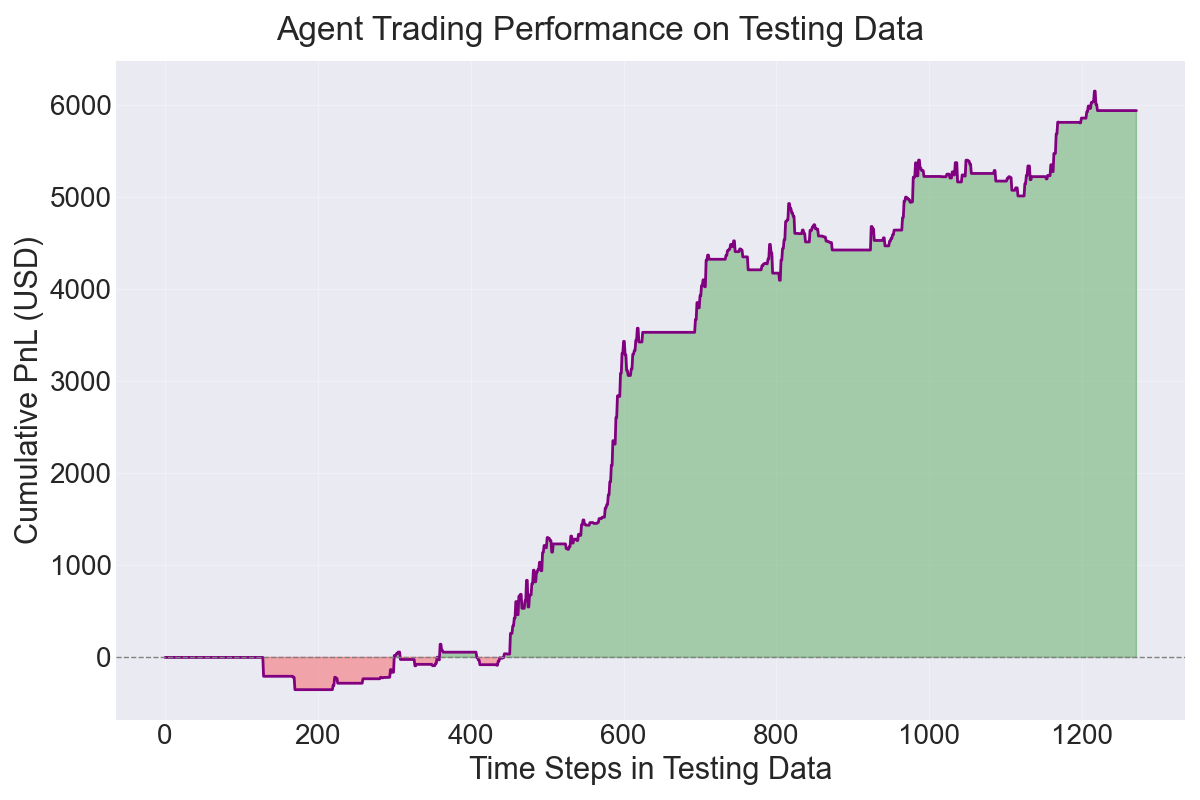}
\caption{Cumulative Profit and Loss (P\&L) of Testing Data}
\label{fig:pnl}
\end{figure}


\begin{table}[htbp]
\centering
\caption{Performance Comparison of A3C and QA3C on the Test Set}
\label{tab:testing-performance}
\renewcommand{\arraystretch}{1.3}
\begin{tabular}{l r r}
\hline
\textbf{Metric} & \textbf{A3C} & \textbf{QA3C} \\
\hline
Total Return & 11.42\% & \textbf{11.87\%} \\
Max Drawdown & 0.92\% & 0.92\% \\
Total Trades & 230 & 231 \\
Win Rate & 54.78\% & \textbf{56.71\%} \\
Best Trade & 0.61\% & 0.61\% \\
Worst Trade & -0.88\% & -0.88\% \\
\hline
\end{tabular}
\end{table}

\begin{table}[htbp]
\centering
\caption{Comparison of 5-Year Total Return: User vs. Major Currency ETFs}
\label{tab:etf-5y-comparison}
\renewcommand{\arraystretch}{1.3}
\begin{tabular}{l p{4cm} r}
\hline
\textbf{Ticker} & \textbf{Name} & \textbf{5-Year Total Return} \\
\hline
FXB & Invesco CurrencyShares British Pound Sterling Trust & 13.81\% \\
FXF & Invesco CurrencyShares Swiss Franc Trust & 12.21\% \\
-- & \textbf{Our Agent} & \textbf{11.87\%} \\
USDU & WisdomTree Bloomberg U.S. Dollar Bullish Fund & 4.91\% \\
FXE & Invesco CurrencyShares Euro Trust & 4.25\% \\
FXC & Invesco CurrencyShares Canadian Dollar Trust & 1.74\% \\
FXA & Invesco CurrencyShares Australian Dollar Trust & -5.79\% \\
FXY & Invesco CurrencyShares Japanese Yen Trust & -29.60\% \\
\hline
\end{tabular}
\end{table}

Figure~\ref{fig:pnl} shows the PnL trajectory: initial flat/mild dip ($-$\$250, steps 0-450) from sidelined/small-loss trades; surge to \$3500 (steps 600-700) capturing USD rally; gradual gains to $>$\$5500 with shallow plateaus, green profits dominating red drawdowns. This confirms compounding small wins with tight risk control.

We also compare the agent's performance to several baseline benchmarks, with a specific focus on the 5-year total returns of major currency ETFs. Table~\ref{tab:etf-5y-comparison} presents a comparison of our agent's 5-year return against those of well-known, large-scale foreign exchange ETFs. Compared to these major currency ETFs, our agent's 11.87\% return in the USD/TWD market is quite competitive. It outperformed pure USD exposure (USDU 4.9\%) and pure Euro or CAD exposure, and also beat AUD and JPY (which had negative returns). It was lower than GBP and CHF, with strong 5-year performances around 12-14\%. However, one must note that the agent achieved its return with far lower drawdown. Many of these currency ETFs likely experienced double-digit percentage drawdowns at times (e.g., the yen ETF had a substantial negative return overall, implying large drawdowns). Our agent's maximum drawdown of 0.92\% is orders of magnitude smaller, indicating a significantly better risk-adjusted return. Statistical t-tests reveal significance (p $<$ 0.05) vs. negative performers. Overall, the QLSTM-QA3C agent learned a profitable strategy, yielding returns that matched or surpassed those of ETFs at a lower risk. QLSTM trends aided decisions, aligning with LSTM-RL hybrids but with quantum edge (not directly compared here).


\section{Discussion}
\begin{figure}[htbp]
\centering
\includegraphics[width=0.49\textwidth]{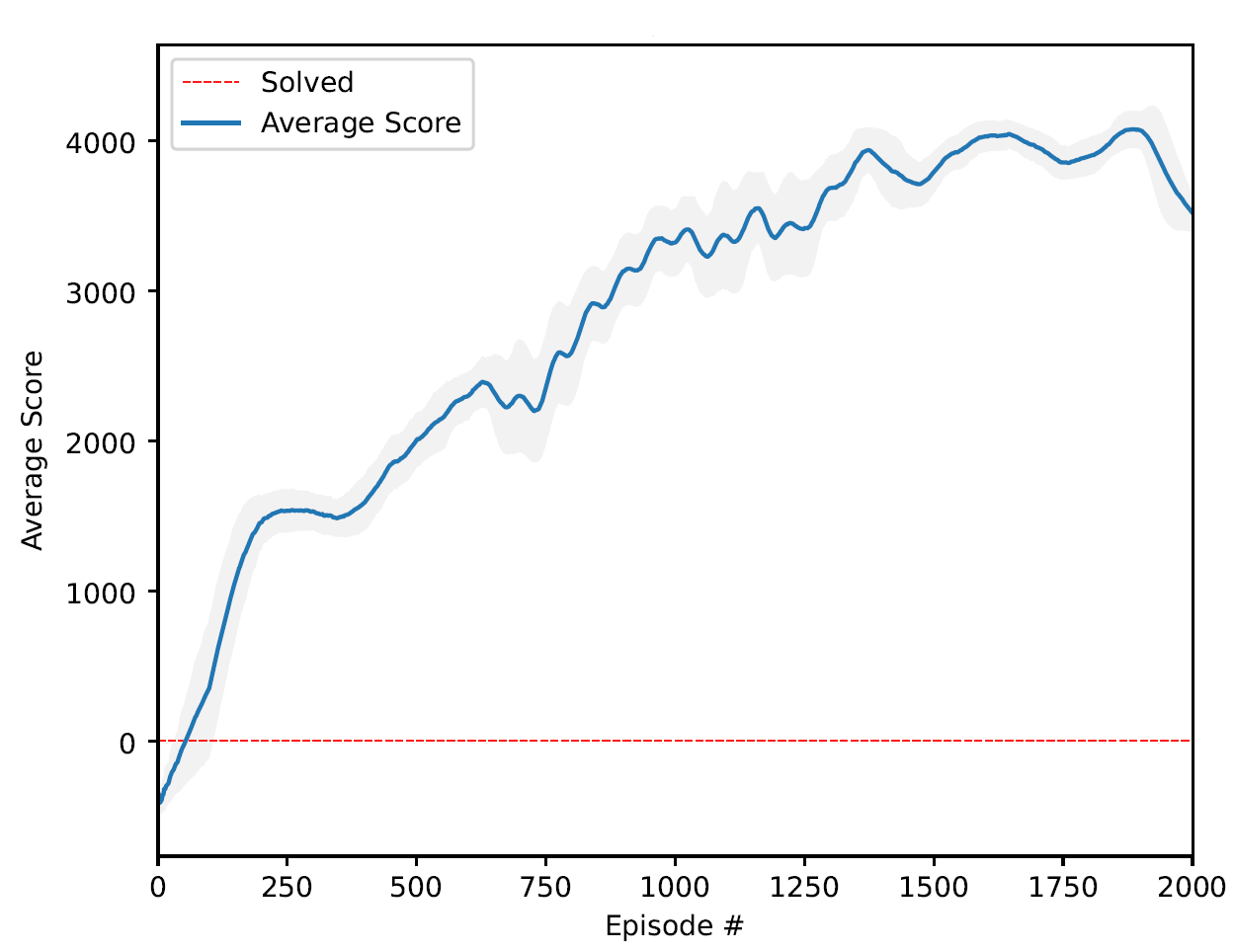}
\caption{Training Reward Convergence Curve}
\label{fig:reward}
\end{figure}

Our experiments demonstrated that reward structures with heavy penalties hinder exploration, causing agents to default to “hold” and stagnate at near-zero rewards. In contrast, our reward design offers a fixed $+10$ bonus for profitable exits and mild penalties ($-2$) for losses, encouraging momentum-following trades. As Figure~\ref{fig:reward} illustrates, the mean training reward rises sharply within 200 episodes, plateaus briefly, then steadily improves, stabilizing around episode 1400—indicating robust convergence.

By linking most rewards to action quality rather than raw profit, the agent learns to take consistent small gains without obsessing over size, maintaining informative gradients even in low-volatility periods. This asymmetric ($+10/-2$) scheme balances optimism and penalty, preventing collapse in exploration.

Regarding the quantum aspect, while current experiments utilize classical simulations, QLSTM could benefit from quantum hardware by more accurately modeling complex dependencies in financial time series through entangled states. Our work contributes to quantum machine learning literature by demonstrating quantum RNNs in finance, and future end-to-end quantum reinforcement learning with variational quantum circuits could further enhance performance.

QLSTM enhances state representations in QA3C, enabling efficient learning and disciplined, momentum-aligned trading with controlled drawdowns. This highlights QML's potential in broader financial areas like stock portfolios and risk management. Limitations include a long-only strategy, no transaction cost modeling, and lack of real-time adaptation. Future work could explore bidirectional trading and quantum hardware deployment.

\section{Conclusion}
In conclusion, this study presents a novel hybrid trading agent that integrates Quantum Long Short-Term Memory (QLSTM) for trend forecasting with Quantum Asynchronous Advantage Actor-Critic (QA3C) for decision-making, demonstrating the potential of quantum-enhanced machine learning in algorithmic trading. Trained on USD/TWD data from 2000-2020 and evaluated on 2020-2025, the QLSTM-QA3C agent delivers 11.87\% total return with a 0.92\% maximum drawdown, outperforming or matching major currency ETFs risk-adjusted while executing 231 disciplined trades marked by momentum-aligned entries, rapid profit realization, and stringent loss mitigation.

Our domain-specific reward function, which balances time decay penalties, asymmetric profit bonuses, and quadratic loss punishments, fosters a high-win-rate, low-volatility strategy resilient to noise. Simulated quantum-inspired models demonstrate practical value in deep reinforcement learning for finance, bridging theoretical quantum machine learning and real-world applications. This extends beyond FX pairs like USD/TWD to portfolio management, derivatives pricing, and risk analysis. In emerging 6G networks, it enables quantum reinforcement learning for resource allocation, distributed sensing, and secure communication. As hardware advances, such architectures may transform trading and quantum networking through improved prediction in volatile markets.

\bibliographystyle{IEEEtran}
\bibliography{bib/qt}

\end{document}